\documentclass[letterpaper]{article} 
\usepackage{aaai23}  
\usepackage{times}  
\usepackage{helvet}  
\usepackage{courier}  
\usepackage[hyphens]{url}  
\usepackage{graphicx} 
\usepackage{natbib}  
\usepackage{caption} 
\usepackage{subcaption}
\usepackage{algorithm}
\usepackage{algorithmic}
\usepackage{amsmath}
\usepackage{amssymb}
\usepackage{breqn}
\usepackage{xcolor}
\usepackage{newfloat}
\usepackage{listings}
\DeclareCaptionStyle{ruled}{labelfont=normalfont,labelsep=colon,strut=off} 
\floatstyle{ruled}
\newfloat{listing}{tb}{lst}{}
\floatname{listing}{Listing}
\title{Deep Reinforcement Learning for Cyber System Defense under \\ Dynamic Adversarial Uncertainties}
\author{
    Ashutosh Dutta\textsuperscript{\rm 1},
    Samrat Chatterjee\textsuperscript{\rm 1}, 
    Arnab Bhattacharya\textsuperscript{\rm 1},     
    Mahantesh Halappanavar\textsuperscript{\rm 1} 
}
\affiliations{
    \textsuperscript{\rm 1}Pacific Northwest National Laboratory\\

    902 Battelle Boulevard\\
    Richland, Washington 99354, USA\\
}

\usepackage{bibentry}

\begin{document}
\maketitle

\begin{abstract}
Development of autonomous cyber system defense strategies and action recommendations in the real-world is challenging and includes characterizing system state uncertainties and attack-defense dynamics. We propose a data-driven deep reinforcement learning (DRL) framework to learn proactive, context-aware, defense countermeasures that dynamically adapt to evolving adversarial behaviors while minimizing loss of cyber system operations. A dynamic defense optimization problem is formulated with multiple protective postures against different types of adversaries with varying levels of skill and persistence. A custom simulation environment was developed and experiments were devised to systematically evaluate the performance of four model-free DRL algorithms against realistic, multi-stage attack sequences. Our results suggest the efficacy of DRL algorithms for proactive cyber defense under multi-stage attack profiles and system uncertainties.
\end{abstract}

\section{Introduction}\label{sec:introduction}
Decision support for cyber system defense in real-world dynamic environments is a challenging research problem that includes the dynamical characterization of uncertainties in the system state and the incorporation of dynamics between attackers and defenders. Cyber defenders typically operate under resource constraints (e.g., labor, time, cost) and must update their strategies and tactics dynamically as the system evolves with/without attack influence from adaptive adversaries. Often, the defender may be unaware of system compromise and must adopt proactive strategies to maintain mission-critical operations. The past decade has seen a growing body of literature focused on the application of game-theoretic approaches for cybersecurity~\cite{roy2010survey, liang2012game}. These approaches typically involve resource allocation optimization with Markov decision process (MDP) models in non-cooperative settings. Cyber system attack and defense modeling methods also include: 1) probabilistic approaches for system reliability and attack outcome dependency~\cite{hu2017multiple, chen2018defending}; 2) Bayesian networks~\cite{frigault2008measuring, shin2015development}; and 3) fault/decision trees~\cite{fovino2009integrating}. Although attack graph-based methods~\cite{ poolsappasit2011dynamic} for identifying system vulnerabilities and known exploits are valuable, within realistic settings, insufficient and imperfect information about system properties and attack goals are typically available to the defender.

Recent advances in reinforcement learning (RL) approaches have led to the development of partially observable stochastic games (POSG) in partial information settings~\cite{macdermed2011markov, sutton2018reinforcement}. Also, cyber system state-space modeling has generated interest in potential use of POSGs for cybersecurity problems~\cite{ramuhalli2013towards, chatterjee2016game, tipireddy2017agent}. However, POSGs tend to be general formulations and are often intractable problems. 
In addition, the state-of-the-art in cyber decision-support also includes the use of partially observable Markov decision process (POMDP) models, as well as distributed POMDPs, for solving a variety of problems such as: 1) cyber risk assessment~\cite{carin2008cybersecurity}, 2) uncertainty in penetration testing~\cite{sarraute2012pomdps}, 3) network task time allocation~\cite{caulfield2015optimizing}, 4) data exfiltration detection~\cite{mc2016data}, and 5) effective deception resource allocation~\cite{islam2021chimera}. Most of these problems either focus only on specific attacks (or attack types), consider a limited defender action space, and assume that a system model is available. 
Recent work in deep RL (DRL) methods for cybersecurity primarily focus either on optimizing network operations or cyber defense against different threat types~\cite{nguyen2019deep,dutta2021constraints}. While progress mostly includes focus on model-free DRL methods, there is a critical need for investigating further the role of different DRL algorithms for cyber defense trained under diverse adversarial settings.     

\begin{figure*}[!ht]
    \centering
    \includegraphics[width=0.9\textwidth]{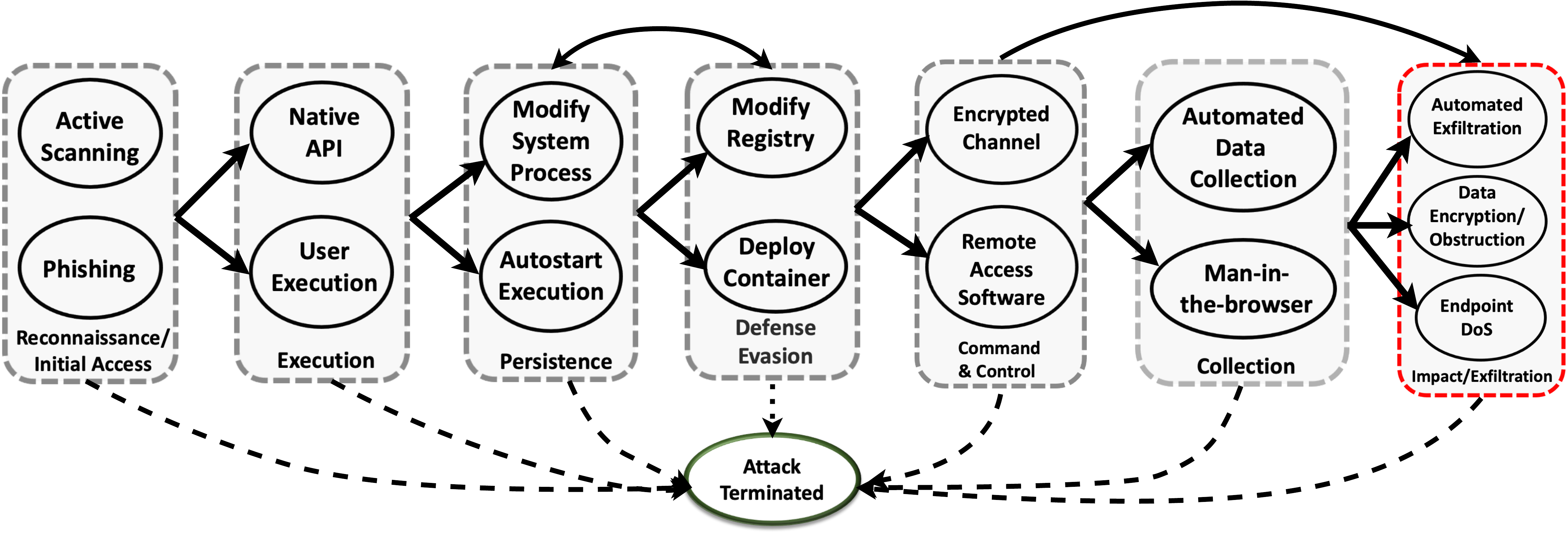}
    \caption{Multi-stage attack propagation represented with MITRE ATT\&CK Tactics and Techniques. (Note: A directed edge between an attack tactic and technique specifies that the attacker may try to implement that technique next after achieving the objective of the attack tactic. Bidirectional arrow represents that \textit{Defense Evasion} can come before \textit{Persistence}.)}
    \label{fig:attack_prog}
\end{figure*}

This paper focuses on the applicability of DRL in optimizing cybersecurity defense planning against strategic multi-stage adversaries. Specifically, the objective of a DRL defense agent is to compute context-aware defense plans by learning network and multi-stage attack behaviors while minimizing impacts to benign system operations. Generally, the dynamics of cyber systems depend on many correlated factors (e.g., traffic volume, network utilization) which may exhibit uncertain behaviors across time due to unanticipated physical link failures, sensor errors, and others. On the other hand, a stealthy cyber adversary may adapt their strategies (i.e., tactics and techniques) based on current network conditions and ongoing attack impact. For example, multi-stage attacks can propagate through the deployment of multiple software processes (e.g., Application Programming Interface (API) calls) to execute attack actions without inducing suspicion, instead of relying on a single process. However, a defender may not always be able to block a process due to its role in maintaining critical operations governed by interdependent processes. Thus, defense planning needs to consider the aggregated impact of different attack actions at different stages of its propagation.  Figure~\ref{fig:attack_prog} presents a multi-stage attack propagation template based on MITRE ATT\&CK framework~\cite{mitre_attack} that was leveraged in our study for evaluating multiple DRL-based defense mechanisms. Within an attack-defense interaction, an adversary can start from any technique of the \textit{Reconnaissance/Initial Access} tactic, and wins if they reach any technique of \textit{Impact/Exfiltration} tactic. Based on defense actions, the attacker may abort (i.e., move to the \textit{Attack Terminated} state or defender's win) or persist to move on to the next stage. 

The main contribution of this paper is to systematically evaluate the performance of multiple DRL algorithms for cyber defense that were trained under diverse adversarial uncertainties. The next section briefly describes DRL algorithms used in this study. Next, we describe the proposed autonomous cyber-defense framework and custom simulation environment. This is followed by experimental results and discussion, and some concluding remarks.      


\section{Background}
\label{sec:background}
\label{subsubsec:rl_policy}
Cyber defense DRL agents compute a policy that recommends optimal action at the current cyber-network state. At time-sequence $t$, the agent executes an action $a_t$ at current state $s_t$ and receives reward $R(s_t,a_t)$ that is used to update the policy. One of the objective of DRL agent is to balance the trade-off between exploration (i.e., executing random actions to understand consequences) and exploitation (i.e., executing optimal actions based on previous exploration knowledge). This research uses $\epsilon$-greedy approaches, where it executes random actions with probability $\epsilon$. Notably, our $\epsilon$ decays with the passage of time.
In this research, we evaluate four different DRL approaches:
\subsubsection{Deep Q-Network (DQN)}Deep Q-Network is a model-free DRL approach that maximizes the Bellman equation:
    \begin{equation} \label{eqn:q_func}
    \begin{split}
        Q_t(s_t,a_t) & = Q_t(s_t,a_t) + \alpha(R_t(s_t,a_t)+ \\
        & \gamma\max_{a}Q_t(s_{t+1},a)-Q_t(s_t,a_t))
    \end{split}
    \end{equation}
where, $Q_t(s_t,a_t)$ is the approximate Q-value (i.e., expected accumulated reward), $R_t(s_t,a_t)$ is the reward ($r_t$) at $t$, and $\alpha$ is the learning rate. At each $t$, the agent executes an action and stores the $(s_t,a_t,r_t,s_{t+1})$ in a replay buffer. After every $n$ steps, the agent calculates the loss using the Eqn. \ref{eqn:q_func} for random batches from the buffer. Here, we use a neural network as a universal function approximator 
of the Q-function $Q_t(s_t, a_t)$~\cite{riedmiller2005neural, sutton2018reinforcement}. Consequently, the neural network parameters are updated by minimizing the loss in Eqn.~\ref{eqn:q_loss} via stochastic gradient descent. Extensions include the Double DQN method that predicts $Q_t(s_{t+1},a)$ and $Q_t(s_{t},a)$ with two different neural networks for more stable Q-function updates. 
        \begin{equation}
            \label{eqn:q_loss}
            Loss = R_t(s_t,a_t)+\gamma \max_{a}Q_t(s_{t+1},a)-Q_t(s_t,a_t).
        \end{equation}
\subsubsection{Actor-Critic} Actor-critic approaches combine both the policy and value iteration methods using the following components: 
\begin{itemize}
\item \textit{Critic} is responsible for policy evaluation and uses a deep neural network (DNN) to estimate the Q-value. Based on the loss in Eqn. \ref{eqn:q_loss}, the critic updates the parameters of the DNN and sends the computed gradients to the actor.
\item \textit{Actor} recommends the optimal action for the current state of a critic using a DNN, whose parameters are updated based on gradients received from the critic. Specifically, the actor searches for the optimal parameters of the DNN, $\theta^*$ (i.e., weights of DNN), that maximize the expected accumulated reward~\cite{sutton2018reinforcement}:
        \begin{equation}
            \label{eqn:policy_iteration}
            J(\pi_{\theta}) = E_{\pi_{\theta}}[\sum_{t=1}^TG(s_t,a_t)]
        \end{equation}
        where, $T$ is the number of decision epochs, and $G(s_t,a_t)$ is the total accumulated reward. The DNN parameters are updated using the following equation:
        \begin{equation}
            \label{eqn:parameter_update}
            \theta = \theta + \alpha\nabla J(\theta)
        \end{equation}
        where, $\alpha$ is the learning rate. In Eqn. \ref{eqn:parameter_update}, $\nabla J(\theta)$ is the gradient descent derived as follows~\cite{sutton2018reinforcement}:
        \begin{equation}
            \label{eqn:gradient_desc}
        \begin{split}
            \nabla J(\pi_{\theta})& =\nabla E_{\pi_{\theta}}[\sum_{t=1}^TG(s_t,a_t)] \\
            & = \nabla E_{\pi_{\theta}}(\sum_{t=1}^T\nabla_{\theta}log \pi_{\theta}(a_t|s_t)G(s_t,a_t))
        \end{split}
        \end{equation}
        where, $\pi(a_t|s_t)$ is the probability of taking action $a_t$ in state $s_t$ subject to the current policy parameters $\theta$.
        We implement three different variants of actor-critic algorithms described as follows.
        \begin{itemize}
        \item \textit{Advantage Actor-Critic Approach (A2C)} This method uses an advantage function, $A(s_t,a_t)$, defined as
         \begin{equation}
            \label{eqn:adv_func}
            A_(s_t,a_t) = G(s_t,a_t) - V(s_t)
        \end{equation}
        to replace the $G(s_t,a_t)$ of Eqn. \ref{eqn:gradient_desc}. The advantage function reduces the high variance to make the policy network more stable. Note that $V(s_t)$ in Eqn.~\ref{eqn:adv_func} is the baseline value achieved at $s_t$.
        \item  \textit{Asynchronous Advantage Actor Critic Approach (A3C):} A3C is different from A2C in that the actor updates the policy parameters asynchronously as soon as any gradient update from any critic (and does not wait for all critics to finish).
        \item  \textit{Proximal Policy Optimization (PPO)} This method clips the divergence of new policy when it is out of the region $[1-e,1+e]$ ($e$ is called clip parameter), in order to avoid drastic deviation from an older evaluated policy. This may be beneficial against sensor noises and errors. PPO optimizes the following clipped surrogate objective function, $L^{CLIP}(\theta)$:
        \begin{equation}\label{eqn:surrogate_objective}
        \begin{split}
            L_t^{CLIP}(\theta) & = \mathbb{E}[\min          
            \{r_t(\theta)A_t,\\
                              & \textrm{clip}\left(\frac{\pi_{\theta}(a_t|s_t)}{\pi_{old}(a_t|s_t)},1-e,1+e\right)A_t\}]
        \end{split}
        \end{equation}
        where, $\pi_{\theta}$ and $\pi_{old}$ represents new and old stochastic policies respectively.
        \end{itemize}
\end{itemize}  

\section{Autonomous Cyber Defense Framework}
\label{sec:framework}
\begin{figure}[!hb]
    \centering
    \includegraphics[scale=0.25]{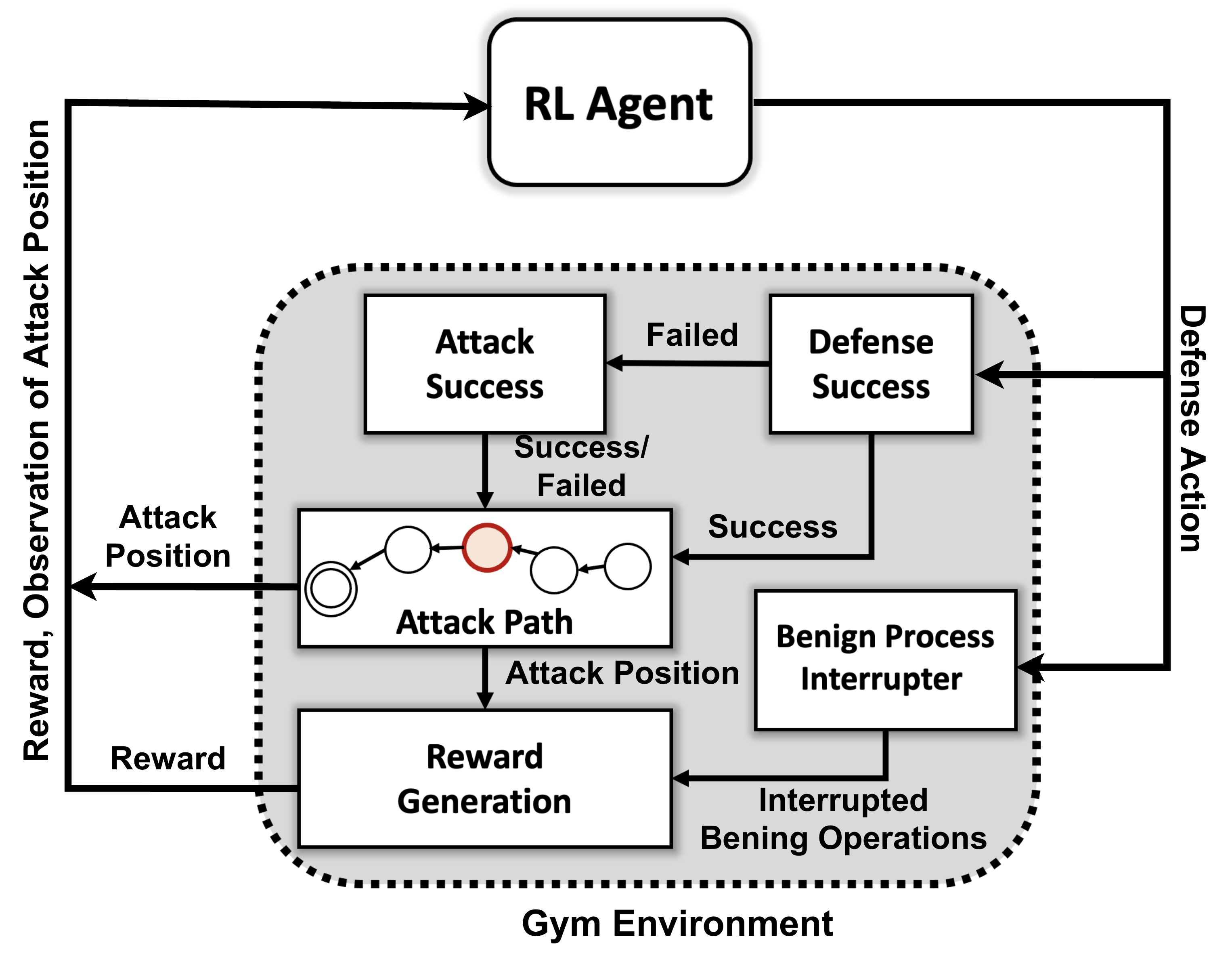}
    \caption{Autonomus cyber defense framework.}
    \label{fig:gym}
\end{figure}

Figure~\ref{fig:gym} presents our proposed autonomous cyber defense framework. The core element of our framework is a custom OpenAI Gym~\cite{brockman2016openai} simulation environment that we developed, where at each time-sequence, a DRL defense agent executes a \textit{Defense Action} and observes \textit{Attack Position} and \textit{Reward} as feedback. Each episode (i.e., multiple time-sequences) considers an attack path (per Figure~\ref{fig:attack_prog}), where the red node in figure~\ref{fig:gym} represents the current \textit{Attack Position}. Next, we describe adversary and defense models in our framework. 

\subsubsection{Adversary Model}
\label{subsubsec:attack_model}
The objective of an adversary is to move from the \textit{Reconnaissance/Initial Access} tactic to the goal tactic of \textit{Impact/Exfiltration} sequentially (per Fig. \ref{fig:attack_prog}). The adversary uses a logic-based model to achieve the objective of a specific tactic if they successfully execute one of their techniques. For example, from attack initiation an adversary may reach the \textit{Reconnaissance/Initial Access} tactic by successfully executing either \textit{Active Scanning} (e.g., scans to find vulnerable machines or services) or \textit{Phishing} (e.g., sends malicious links to users). At any time, the attack position is the last successfully executed attack technique. The attack graph in Fig. \ref{fig:attack_prog} satisfies the monotonicity property, which means that an adversary never discards a specific attack position once they successfully execute it.

There are many possible chains/paths to move from \textit{Reconnaissance/Initial Access} position to \textit{Impact/Exfiltration}, from which, the attacker can choose any path based on the attack strategy. We assume that an adversary can change their strategy according to observations of previous attack impacts and network/system conditions. As a result, they can evade or defeat any static defense policy by discovering the deployed countermeasures. 

To execute a technique successfully, the adversary needs to exploit at least one of the existing vulnerabilities that are still not discovered or cannot be patched due to usability, safety, or stability requirements. To do so, the adversary may follow diverse procedures based on the following logic. An attack attempt fails if they cannot exploit any relevant vulnerability due to countermeasures or lack of expertise. From any tactic/technique, the adversary moves to \textit{Attack Terminated} position if their multiple attempts to execute next attack technique fails due to successful defense actions. Hence, there are two attack consequences: (i) adversary wins if they can successfully execute any technique under the \textit{Impact/Exfiltration} position; (ii) adversary loses if they move to  the \textit{Attack Terminated} position. It is important to note that the adversary's strategy is implicitly captured in the description of the system's state evolution.

\subsubsection{Defense Model}
\label{subsubsec:defense_model}
The main objective of the defender (DRL agent) is to proactively prevent the adversary to reach the Impact/Exfiltration tactic phase, while minimizing loss due to interrupting benign operations. 
To determine the optimal defense, the agent needs to infer the current attack position and predict the next attack action. However, lack of domain-specific information on system complexities and adversarial behavior results in the defender facing the following uncertainties that prevents the creation of an apriori system dynamics model:
\begin{itemize}
    \itemsep 0em
    \item \textit{Uncertain Next Attack Technique:} The defense agent cannot predict the next attack technique due to two reasons. First, the defense agent has no prior knowledge of the attack graph in Fig. \ref{fig:attack_prog} as it requires domain-specific real-world attack sequences that are hard to obtain. Second, an adversary does not always follow the same strategy (attack sequence). For example, after \textit{User Execution}, an adversary can either try to \textit{Modify Registry} or \textit{Modify System Process}. 
    
    \item \textit{Uncertain Next Attack Procedure:} The defense agent does not know the next attack procedure (i.e., attack actions required to implement an attack technique) or its likelihood due to lack of domain data. 
    
    \item \textit{Imperfect and Incomplete Observations:} Deployed alert mechanisms may not prove the current position of the adversary due to two reasons: (1) limited observability of processes, and (2) uncertain mapping from observations to attack techniques.  
\end{itemize}


For our DRL experimentation, we assume that an adversary is at an initial access position at the start of each episode, and the defender leverages alert systems that monitor API calls to understand the current attack position. As mentioned before, this alert information is not only incomplete due to not observing all API calls but also imperfect due to probability of errors in mapping aggregated API calls to MITRE tactic and technique.
Moreover, the defender only partially knows attack position but does not know which process is malicious, and therefore, seeks to learn the adversary's dynamic attack strategy governed by attack action sequences. 

\section{Defense Optimization Problem}
\label{sec:form_df_model}
We formulate the cyber defender's optimization problem using a Sequential Decision Process (SDP)~\cite{braziunas2003pomdp} construct, where the next attack technique depends on the current attack position and defense action.
Note here that the adversary's optimal strategy is not learned in this work; instead, we train the defense agent against different adversarial strategies corresponding to distinct attack behaviors. Our SDP model is a tuple with four parameters: $(S,A,R,\gamma)$, where $S$ is the cyber system state space, $A$ is the defense action space, $R$ is the reward function, and $\gamma$ is the discount factor. The agent only knows about $S$, $A$, and $\gamma$ which define its interaction with the environment. Though the agent is unaware of the mathematical form of $R$. The objective in the SDP model is to compute an optimal action, $a^*\in A$, for any current state, $s\in S$ that maximizes the cumulative reward over a finite attack horizon.

At the start of each time $t$, the agent executes a defense action, $a_t\in A$, based on current environment state, $s_t\in S$, and receives feedback from the environment. Such feedback consists of current attack position, $s_{t+1}\in S$, as observation and defense payoff, $r_t$, as a reward. The payoff of $a_t$ depends on its effectiveness on whether $a_t$ prevented the next attack technique or not. To clarify, $a_t$ is an effective defense action at $s_t$ if it forces the attacker to stay at the previous position ($s_{t+1}=s_t$) or move to \textit{Attack Terminated} node.  Thus, this model also implicitly integrates the expected attack behavior into decision-model through learning effectiveness of defense actions at particular environment condition. 
Next, we describe the optimization model elements in detail.

\subsubsection{State Space ($S$)}
\label{subsubsec:state}
Our state space, $S$, consists of 17 states, where each state, $s\in S$, is a sparse vector that represents an unique attack position. Hence, the current state at any time $t$ specifies the position of adversary at $t$, based on which, the defense agent aims to choose the optimal defense action for that time-sequence. We consider three types of states: (1) 15 attack-technique states corresponding to each attack technique in Fig. \ref{fig:attack_prog}; (2) \textit{Attack Initiated} state, and (3) \textit{Attack Terminated} state.  
The initial state is the \textit{Attack Initiated} state, and the adversary moves to \textit{Attack Terminated} state if they abort the attack due to failure or detection. Moreover, the adversary moves to a new state by successfully executing the associated attack technique. We consider two types of goal states:
\begin{itemize}
    \item \textit{Attack Goal State:} Adversary wins if they reach (1) \textit{Automated Exfiltration}, (2) \textit{Data Encryption/Obstruction}, or (3) \textit{Endpoint Denial of Service (DoS)} state of \textit{Impact/Exfiltration} tactic;
    \item \textit{Defense Goal State:} Defender wins if the adversary moves to \textit{Attack Terminated} state.
\end{itemize}


\subsubsection{Defense Action Space ($A$)}
\label{subsubsec:defense_action}
The defender's action space, $A$, has three different modes of operation:
\begin{enumerate}
    \itemsep0em 
    \item \textit{Inactive}: The defender remains silent and does nothing;
    \item \textit{Reactive}: The defender removes all processes that called or executed actions related to current attack position or last attack action;
    \item \textit{Proactive}: The defender blocks a specific set of API calls or operations to prevent the next attack action.
\end{enumerate}

The defender remains \textit{Inactive} for strategic reasoning or to avoid termination of critical benign processes. Whereas, for both \textit{reactive} and \textit{proactive} defense approaches, benign operations may be interrupted. We consider only one reactive action assuming that it can remove all processes associated with last attack action (whether successful or not). For proactive defense, we consider 21 distinct defense actions that block unique sets of API calls and operations or adopt particular methods/measurements to prevent specific attack techniques. For instance, a proactive defense action: Restrict Registry Permission blocks the ability to change certain keys to prevent adversary’s autostart execution (\textit{Persistence} tactic) or registry modification (\textit{Defense Evasion} tactic). Another proactive action such as Restrict File and Directory Permission (only certain set of sensitive files) can also stop autostart execution. However, success likelihood of different defense actions vary in both defense effectiveness (i.e., preventing attack action) and expected false positive rate (i.e., terminating benign operations). Hence, the defense agent must choose the best action considering defense effectiveness and expected false positive rate. Moreover, this research also assumes that the defender can execute one action at a time-sequence. Hence, for proactive defense approach, the defender's decision-model needs to understand what the adversary may do next to execute an effective mitigation action. The defender wins if the adversary moves to \textit{Attack Terminated} position due to failure to execute new attack techniques or being detected and removed.

\subsubsection{Reward Function ($R$)}
\label{subsubsec:reward}

We consider the following reward function for the defense agent:
\begin{equation}\label{eqn:reward_func}
    R=-p_g(s)\times I_g-\mathcal{I}_v\times I_g-C_f
\end{equation}
where, $\mathcal{I}_v=-1$ if the defender wins, and 0 otherwise.
Note that $p_g (s)$ is the probability of the adversary reaching the \textit{Impact/Exfiltration} tactic from state $s$, and $I_g$ is the impact/loss of a successful attack execution. Equation \ref{eqn:reward_func} comprise of three terms: (1) $p_g (s)\times I_g$ quantifies the risk at state $s$ due to the probability ($p_g (s)$) of attacker’s reaching to goal; (2) $I_v\times I_g$ quantifies the penalty or incentive to the defense agent when the adversary wins or loses, respectively; and (3) $C_f$ is the cost of executing any defense action in $A$. The cost $C_f$ depends on the aggregated loss due to interrupting benign operations, and defense implementation or operational cost. We assume that the defense implementation cost is same for all mitigation actions and zero cost for \textit{Inactive} action. 

\section{Experiments}
\label{sec:experiments}
To implement our framework and solve the cyber defense optimization problem, an experimental plan was established to evaluate the performance of four DRL approaches (i.e., DQN, A2C, A3C, and PPO). The experimental setup, training and testing scenarios, adversary types, and simulation environment are described next.

\subsection{Experimental Setup} We designed our experiments using Python 3.7~\cite{van1995python}, and used \textit{RLlib} library~\cite{liang2018rllib} for implementing DRL algorithms. We have simulated our experiments using a Dell Alienware machine with 16-core 3GHz Intel Core i7-5960X processor, 64GB RAM, and three 4GB NVIDIA GM200 graphics cards. Although MITRE ATT\&CK framework contains 11 tactics and many more techniques, we consider 7 tactics and 15 techniques for our experiments. Our defense action space $A$ consists of 23 mitigation actions, including 21 proactive actions. Table \ref{tab:exp_parameter} presents learning parameters that we used for all our experiments. 

\begin{table}[!hb]
\centering
\scalebox{0.8}{
    \begin{tabular}{|c|c|}
        \hline 
        Parameter Name                                                                     & Value            \\ \hline \hline
        Entropy Coefficient                                                                & 0.05             \\ 
        Initial exploration probability                                                    & 1.0              \\ 
        Final exploration probability                                                      & 0.04             \\ 
        Exploration delay period                                                           & 300,000          \\ 
        Number of workers & 4                \\ 
        Rollout fragment length                                                            & 12               \\ 
        Batch size                                                                         & 48               \\ 
        PPO clip value                                                                     & 0.4              \\ 
        Training epochs                                                                    & 100              \\ 
        Steps per training epoch                                                           & 25000            \\ \hline
    \end{tabular}
}
\caption{Parameters used for training and testing simulation experiments.
(Note: \textit{Number of workers} refers to the number of parallel processes, and \textit{Exploration delay period} refers to the number of training steps that decays exploration probability from 1.0 to 0.04.)}
\label{tab:exp_parameter}
\end{table}

\subsection{Training and Testing Scenarios} We generated all possible distinct attack paths from initial attack position (i.e., initial attack state) to any state of the last attack tactic (i.e., \textit{Impact/Exfiltration} tactic). We used 80\% attack paths for training and 20\% attack paths for testing, where each attack propagation path is a unique sequence of attack techniques executed by an adversary to achieve their objective.

During training, each unique episode contains one attack propagation path, which ends if the defender/adversary wins or loses. The adversary wins or the defender loses in an episode if the adversary successfully executes all attack techniques across the attack path to satisfy \textit{Impact/Exfiltration} tactic. However, an adversary loses or defender wins if they move to \textit{Attack Terminated} state due to failing at least a number of times, $n$, across a path. 

\subsection{Adversary Types}\label{sec:attack_type} For our experiments, we determine adversary types based on two attack parameters: (1) \textit{skill}, and (2) \textit{persistence}. With greater attack skill, success rate in executing next attack technique is higher due to increased capability in exploiting vulnerabilities. We define $\rho$ as an attack skill parameter, representing attack success rate in exploiting a vulnerability. We assume that all vulnerabilities have same exploitability (i.e., impact and complexity). On the other hand, greater attack persistence indicates that the adversary does not abort their objective or cannot be readily detected despite their failed attempts. We define $\tau$ as an attack persistence parameter, representing the number of failed attempts before moving to attack terminated state (i.e., defense win). By tuning $\rho$ and $\tau$, we consider three different attack profiles/strategy: (1) Attack profile 1 ($Av_1$): $\rho=0.75$ and $\tau=4$, (2) Attack profile 2 ($Av_2$): $\rho=0.85$ and $\tau=5$, and (3) Attack profile 3 ($Av_3$): $\rho=0.95$ and $\tau=7$. For example, $Av_2 (\rho=0.85,\tau=5)$ indicates that the adversary with profile 2 has 85\% success likelihood in exploiting a specific vulnerability and tolerates maximum 5 failed attempts in an episode. Thus, attack profile 3 represents the most sophisticated adversary, and attack profile 1 represents a na\"ive adversary. It is important to note that the adversary may change attack procedures if their previous attempt in exploiting a specific attack technique fails.

\subsection{Simulation Environment} 
\label{subsubsec:sml_env}
We developed a custom OpenAI Gym simulation environment (not in the public domain at this time) for the autonomous cyber defense framework illustrated in Figure~\ref{fig:gym}. The DRL defense agent has no knowledge of the environment and attack behavior. The defense agent receives observations about next attack position and reward as feedback. The agent then determines the current attack position based on recent observation. We assume that system alerts can be translated to a specific attack position with 85\%, 75\%, and 65\% accuracy against $Av_1$, $Av_2$, and $Av_3$, which indicates that the adversary is more stealthy with increased level of sophistication.

Within the simulation environment, at the start of each episode, an attack propagation path is selected, and the environment is set to the initial attack position. At the start of each timestep, a \textit{Defense Success} module determines whether the current defense action stopped the current attack technique or not, based on the correlations among attack procedures and defense actions. Another module, \textit{Benign Process Interrupter}, sends the list of benign processes/operations interrupted due to the recent defense action to the \textit{Reward Generation} module. We assume a power-law distribution function to generate the number of interrupted benign operations, which gradually becomes lower towards the last attack tactic. If the defense action is successful, the adversary stays at their current position. Otherwise, if the defense action fails, \textit{Attack Success} determines the success of the current attack technique based on attack skill ($\rho$) of attack profile. If the attack is successful, the adversary moves to next position; otherwise, they remain at the current position or move to \textit{Attack Terminated state} (end of episode). 
The reward is generated based on attack position, interrupted benign operations, and others using Eqn. \ref{eqn:reward_func}. At the end of the timestep, reward and next observation are sent to the DRL defense agent that is used to update and refine their policy.

\section{Evaluation}
\label{sec:evaluation}

 \begin{figure*}[!ht]
        \centering
         \begin{subfigure}{0.8\textwidth}
         \centering
        	\includegraphics[scale=0.07]{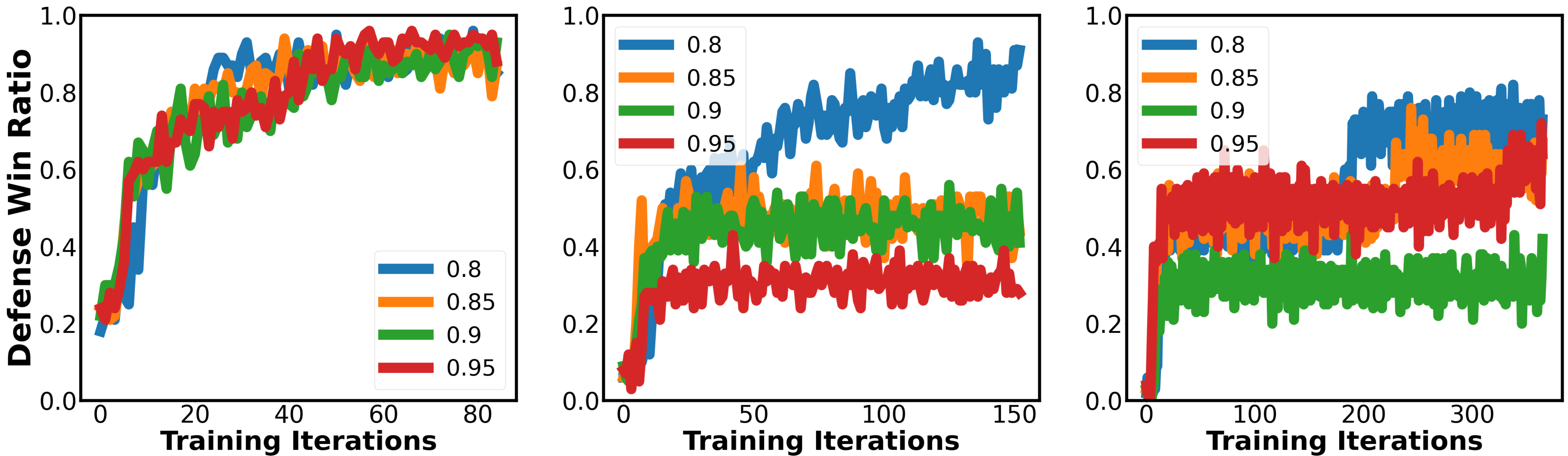}
        	\caption{Convergence for different values of $\gamma$ ($\alpha$ set to 0.005).}
        	\label{fig:eval_tr_a2c}
         \end{subfigure}
         \\
         \begin{subfigure}{0.8\textwidth}
         \centering
          \includegraphics[scale=0.07]{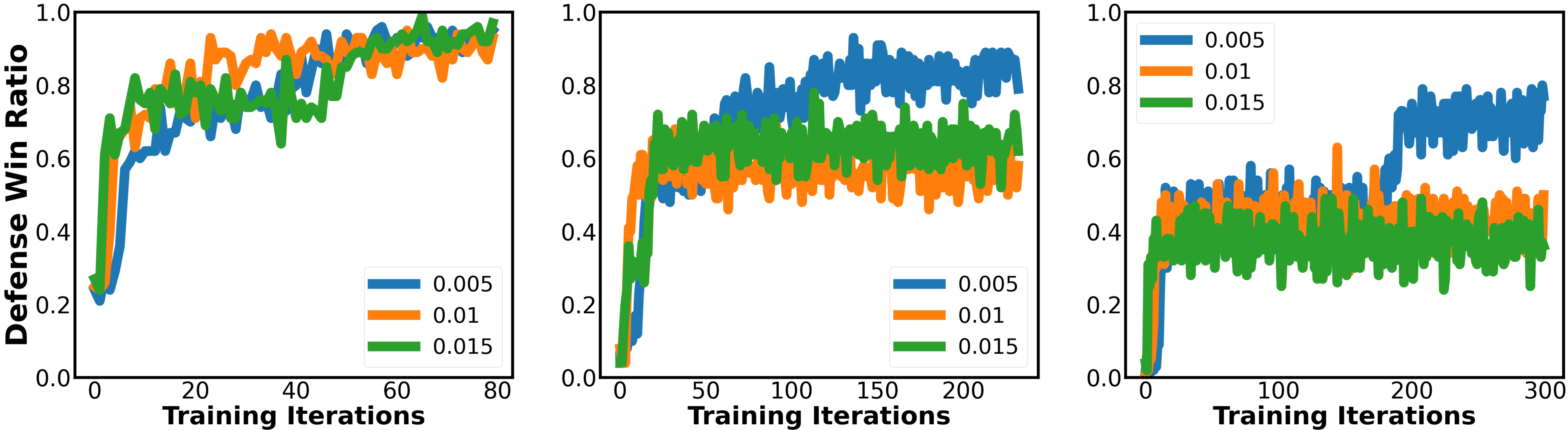}
          \caption{Convergence for different values of $\alpha$ (with $\gamma$ set to optimized value of 0.8).}
          \label{fig:eval_tr_a2c_lr}
         \end{subfigure}
         \caption{Sensitivity of A2C for different values of $\gamma$ and $\alpha$ against attack profiles $Av_1$, $Av_2$, and $Av_3$ (left to right).}
         \label{fig:eval_training}
    \end{figure*}

We assess the performance of different DRL algorithms using the \textit{Defense-Win Ratio} (DWR) metric, which evaluates the fraction of episodes where the defender won in a single batch, where each batch comprises of 200 episodes. Using the DWR ratio, we analyze (i) the convergence of the DRL algorithms during training, and (ii) the performance of the trained defense models against unseen attack sequences during testing. We evaluate all four DRL algorithms against three attack profiles (as described in Section \ref{sec:attack_type}). 

\subsection{Hyperparameter Optimization and Training} 
For brevity, here we only illustrate the training performance of DQN and A2C using the DWR metric by varying two hyper-parameters: discount factor ($\gamma$) and learning rate ($\alpha$). In the following sections, each figure has two rows and three columns; the columns refer to the three attack profiles, while the rows correspond to the hyperparameters (the first row is for $\gamma$ and the second row is for $\alpha$). In this research, for all DRL algorithms, we have used fully connected neural network with 2 hidden layers where each of them has 256 neurons. We have used \textit{tanh} as activation function in all the cases.

\subsubsection{A2C}Figure \ref{fig:eval_tr_a2c} illustrates the sensitivity of A2C to different values of $\gamma$, keeping $\alpha$ fixed to 0.005. Against attack profile 1, all A2C instances converge at DWR value of 0.95 within 40 iterations ($40\times 200$ episodes). Against attack profile 2, the A2C instance with $\gamma=0.8$ achieves the highest DWR ratio, while all other discount factors induce poor performance. For attack profile 3, $\gamma=0.8$ reaches the highest DWR of 0.8 while requiring 200 iterations. Thus, attack sophistication not only increases the convergence time but also reduces the defense agent's success rate. Averaging across all attack profiles, A2C performs best with $\gamma=0.8$, which shows that the defense agent must balance the trade-off between the current reward and possible future payoff. Fig. \ref{fig:eval_tr_a2c_lr} exhibits the sensitivity of A2C (with optimal $\gamma$ set to 0.8) for different values of $\alpha$. We observed that the algorithm performance does not change significantly against attack profile 1. Interestingly, against both attack profiles 2 and 3, A2C with $\alpha=0.0005$ shows the best performance; this is possibly because the algorithms get stuck to local minima for larger values of $\alpha$. Therefore, for A2C we set $\alpha=0.005$ and $\gamma=0.8$ for the test experiments.

\subsubsection{DQN}
\begin{figure*}[!ht]
        \centering
         \begin{subfigure}{\textwidth}
         \centering
        	\includegraphics[scale=0.07]{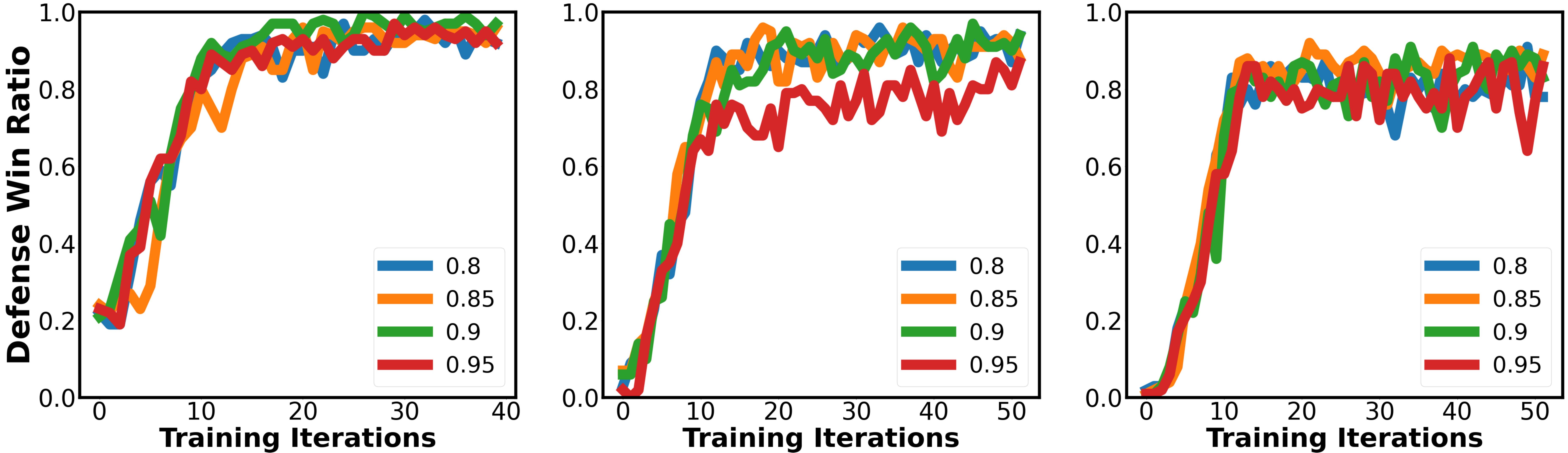}
        	\caption{Convergence for different values of $\gamma$ ($\alpha$ set to 0.005).} \label{fig:eval_dqn_df}
         \end{subfigure}
         \\
         \begin{subfigure}{\textwidth}
         \centering
          \includegraphics[scale=0.07]{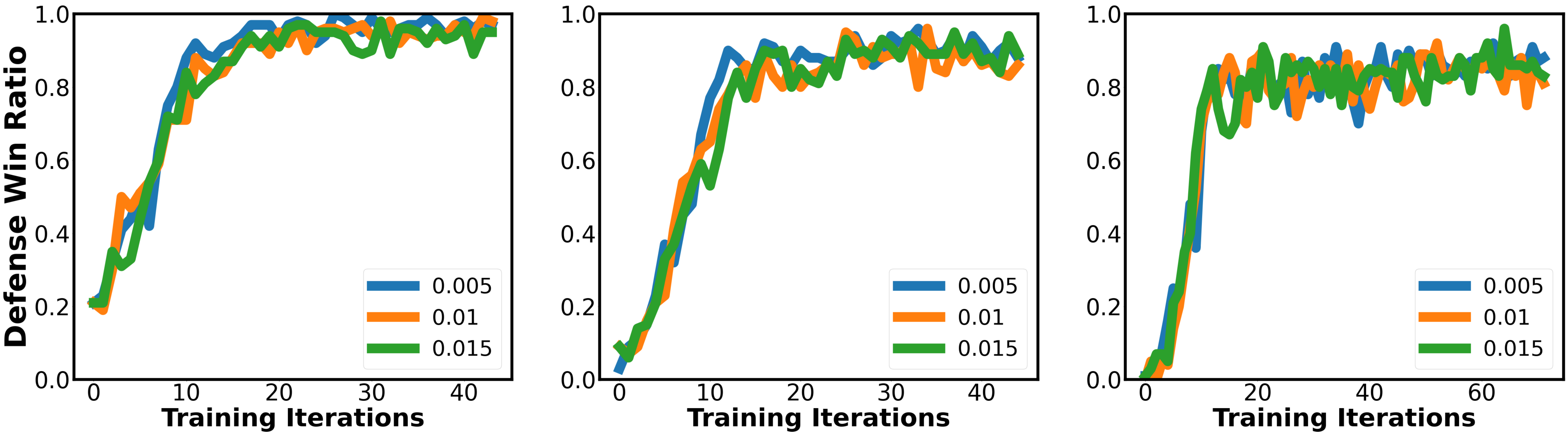}
          \caption{Convergence for different values of $\alpha$ (with $\gamma$ set to optimized value of 0.8).}
          \label{fig:eval_dqn_lr}
         \end{subfigure}
         \caption{Sensitivity of DQN for different values of $\gamma$ and $\alpha$ against attack profiles $Av_1$, $Av_2$, and $Av_3$ (left to right)}
         \label{fig:eval_training}
    \end{figure*}
    We followed the same training approach for DQN as that of A2C. From Fig. \ref{fig:eval_dqn_df}, we observe that DQN converges to optimal policy within 10 iterations, which is much faster compared to A2C; in fact, it was better than all of the DRL algorithms we tested. Moreover, DWR was higher than other DRL methods for each attack profile, which showed the superior performance of DQN for environments with discrete states and actions. Moreover, Fig. \ref{fig:eval_dqn_lr} illustrated that DQNperformance does not change significantly with change in $\alpha$. For testing, we set $\gamma=0.8$ and $\alpha=0.01$ for DQN.

\subsection{Testing Results}
\begin{figure*}[htb!]
\centering
\includegraphics[scale=0.3]{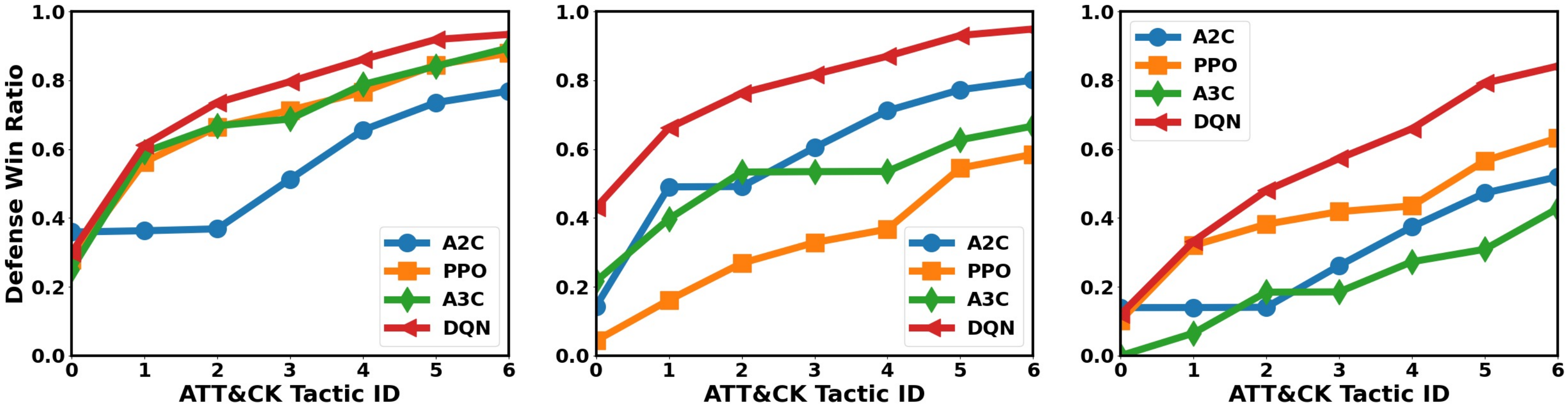}
\caption{DWR performance during testing against the three attack profiles $Av_1$, $Av_2$, and $Av_3$ (from left to right).}
\label{fig:eval_test_one}
\end{figure*}
        
\begin{table*}
\centering
    \begin{tabular}{|c|ccc|ccc|ccc|}
        \hline
        {\begin{tabular}[c]{@{}c@{}}DRL\\ Algorithm\end{tabular}} & \multicolumn{3}{c|}{\begin{tabular}[c]{@{}c@{}}Attack Profile 1\\ (in \%)\end{tabular}}                                                                                                                           & \multicolumn{3}{c|}{\begin{tabular}[c]{@{}c@{}}Attack Profile 2\\ (in \%)\end{tabular}}                                                                                                                           & \multicolumn{3}{c|}{\begin{tabular}[c]{@{}c@{}}Attack Profile 3\\ (in \%)\end{tabular}}                                                                                                                           \\ \cline{2-10} 
                                                                                & \multicolumn{1}{c|}{\begin{tabular}[c]{@{}c@{}}Tactic\\ ID: 3\end{tabular}} & \multicolumn{1}{c|}{\begin{tabular}[c]{@{}c@{}}Tactic\\ ID: 6\end{tabular}} & \begin{tabular}[c]{@{}c@{}}Mean\\ Reward\end{tabular} & \multicolumn{1}{c|}{\begin{tabular}[c]{@{}c@{}}Tactic\\ ID: 3\end{tabular}} & \multicolumn{1}{c|}{\begin{tabular}[c]{@{}c@{}}Tactic\\ ID: 6\end{tabular}} & \begin{tabular}[c]{@{}c@{}}Mean\\ Reward\end{tabular} & \multicolumn{1}{c|}{\begin{tabular}[c]{@{}c@{}}Tactic\\ ID: 3\end{tabular}} & \multicolumn{1}{c|}{\begin{tabular}[c]{@{}c@{}}Tactic\\ ID: 6\end{tabular}} & \begin{tabular}[c]{@{}c@{}}Mean\\ Reward\end{tabular} \\ \hline
        A2C                                                                     & \multicolumn{1}{c|}{51.3}                                                   & \multicolumn{1}{c|}{76.9}                                                   & 56.7                                                  & \multicolumn{1}{c|}{60.5}                                                   & \multicolumn{1}{c|}{80}                                                     & 59.8                                                  & \multicolumn{1}{c|}{26.2}                                                   & \multicolumn{1}{c|}{51.9}                                                   & 27.11                                                 \\ 
        PPO                                                                     & \multicolumn{1}{c|}{71.3}                                                   & \multicolumn{1}{c|}{87.9}                                                   & 70.3                                                  & \multicolumn{1}{c|}{32.9}                                                   & \multicolumn{1}{c|}{58.5}                                                   & 36.8                                                  & \multicolumn{1}{c|}{41.9}                                                   & \multicolumn{1}{c|}{63.2}                                                   & 40                                                    \\ 
        A3C                                                                     & \multicolumn{1}{c|}{68.8}                                                   & \multicolumn{1}{c|}{89.4}                                                   & 72                                                    & \multicolumn{1}{c|}{53.5}                                                   & \multicolumn{1}{c|}{66.7}                                                   & 48                                                    & \multicolumn{1}{c|}{18.5}                                                   & \multicolumn{1}{c|}{42.8}                                                   & 18.5                                                  \\ 
        DQN                                                                     & \multicolumn{1}{c|}{79.6}                                                   & \multicolumn{1}{c|}{93.3}                                                   & 77                                                    & \multicolumn{1}{c|}{82}                                                     & \multicolumn{1}{c|}{95}                                                     & 77.2                                                  & \multicolumn{1}{c|}{57.3}                                                   & \multicolumn{1}{c|}{84.1}                                                   & 60.7                                                  \\ \hline
     \end{tabular}
\caption{Testing performance comparison among different DRL algorithms.}
\label{tab:test_table}
\end{table*}
We discuss the testing performance of the DRL algorithms against the three attack profiles with unseen attack sequences not used during training. Fig.~\ref{fig:eval_test_one} illustrates the cumulative result in defending against the adversary at different phases. Here, each phase corresponds to a distinct adversary tactic. \textit{Tactic ID: 0} specifies the initial attack position before \textit{Reconnaissance}, \textit{Tactic ID: 6} specifies the \textit{Collection}, and all other tactics in Fig. \ref{fig:attack_prog} are numbered sequentially from \textit{Tactic ID: 1} to \textit{Tactic ID: 5}. Note that \textit{Impact/Exfiltration} (tactic ID 7) is not shown in Fig.~\ref{fig:eval_test_one} as the defender loses if the attacker reaches that state. Table~\ref{tab:test_table} reports how many attack sequences were stopped at the corresponding attack tactic. For example, the column corresponding to \textit{Tactic ID: 3} specifies how many attacks were stopped at \textit{Defense Evasion}; similarly for the other columns. The \textit{Mean Reward} column specifies how much reward is achieved compared to the best reward (i.e., when stopping all attacks at Tactic ID: 0). 

The objective of the defense DRL agent is not only to stop the adversary from moving to \textit{Impact/Exfiltration} state but also to stop the attack progression as early as possible. Against all attack profiles, we observed that DQN exhibited the best performance in stopping the adversary as soon as possible. As can be seen in Table~\ref{tab:test_table} and Fig.~\ref{fig:eval_test_one}, DQN stops the three attack profiles within the \textit{Defense Evasion} (tactic ID 3) in 79.6\%, 82\%, and 57.3\% of the cases respectively, while achieving corresponding defense success rates of 93.3\%, 95\%, and 84.1\%. Against $Av_3$, DQN has a lower success rate than other attack profiles, as the sophisticated adversary hardly fails in exploiting a vulnerability. Besides, $Av_3$ is persistent and does not give up easily in spite of failed attempts, which leads to lower DWR values for the defense agent. Note that other DRL algorithms did not consistently perform well against all three attack profiles. In fact, the next-best performing algorithm against $Av_3$ had approximately 50\% success rate. This is possibly due to the fact that actor-critic methods typically require more training samples than DQN to show better test accuracy. 

\section{Conclusion}
\label{sec:conclusion}
Application of DRL methods for cyber system defense are promising, especially under dynamic adversarial uncertainties and limited system state information. Evaluating multiple DRL algorithms trained under diverse adversarial settings is an important step toward practical autonomous cyber defense solutions. Our experiments suggest that model-free DRL algorithms can be effectively trained under multi-stage attack profiles with different skill and persistence levels, yielding favorable defense outcomes in contested settings. However, some practical challenges that need to be addressed further in using model-free DRL include~\cite{dulac2019challenges}: (i) explainability of the black-box DRL policies, (ii) vulnerability to adversarial noise and data poisoning, and (iii) convergence for large state-action spaces. Future work will include developing DRL-based transfer learning approaches within dynamic environments for distributed multi-agent defense systems.      


\section{Acknowledgments}
This research was supported by the U.S. Department of Energy, through the Office of Advanced Scientific Computing Research's “Data-Driven Decision Control for Complex Systems (DnC2S)” project. Part of this research was supported by the Mathematics for Artificial Reasoning in Science (MARS) initiative at  Pacific Northwest National Laboratory (PNNL) under the Laboratory Directed Research and Development (LDRD) program. PNNL is multiprogram national laboratory operated by Battelle for the U.S. Department of Energy under contract DE-AC05-76RL01830.

\bibliography{aaai23}
\end{document}